\documentclass[10pt, a4paper]{article}
\usepackage{lrec2022} 
\usepackage{multibib}
\newcites{languageresource}{Language Resources}
\usepackage{graphicx}
\usepackage{tabularx}
\usepackage{soul}
\usepackage{titlesec}
\titleformat{\section}{\normalfont\large\bfseries\center}{\thesection.}{1em}{}
\titleformat{\subsection}{\normalfont\SmallTitleFont\bfseries\raggedright}{\thesubsection.}{1em}{}
\titleformat{\subsubsection}{\normalfont\normalsize\bfseries\raggedright}{\thesubsubsection.}{1em}{}
\renewcommand\thesection{\arabic{section}}
\renewcommand\thesubsection{\thesection.\arabic{subsection}}
\renewcommand\thesubsubsection{\thesubsection.\arabic{subsubsection}}

\usepackage{epstopdf}
\usepackage[utf8]{inputenc}

\usepackage[hidelinks]{hyperref}
\usepackage{xstring}

\usepackage{color}

\usepackage{booktabs}
\usepackage{adjustbox}
\usepackage[para]{threeparttable}
\usepackage{lipsum}
\usepackage{float}
\usepackage{multirow}

\title{AnnoTheia: A Semi-Automatic Annotation Toolkit\\for Audio-Visual Speech Technologies}

\name{José-M. Acosta-Triana\textsuperscript{1}, David Gimeno-Gómez\textsuperscript{2}, Carlos-D. Martínez-Hinarejos\textsuperscript{2}} 

\address{\textsuperscript{1} ValgrAI - Valencian Graduate School and Research Network of Artificial Intelligence\\ Camino de Vera, s/n, 3Q Building, 46022, València, Spain\\ \textsuperscript{2} Pattern Recognition and Human Language Technologies Research Center\\ Universitat Politècnica de València, Camino de Vera, s/n, 46022, València, Spain \\ {\tt joactr@inf.upv.es, dagigo1@dsic.upv.es, cmartine@dsic.upv.es}         
}

\abstract{
More than 7,000 known languages are spoken around the world. However, due to the lack of annotated resources, only a small fraction of them are currently covered by speech technologies. Albeit self-supervised speech representations, recent massive speech corpora collections, as well as the organization of challenges, have alleviated this inequality, most studies are mainly benchmarked on English. This situation is aggravated when tasks involving both acoustic and visual speech modalities are addressed. In order to promote research on low-resource languages for audio-visual speech technologies, we present AnnoTheia, a semi-automatic annotation toolkit that detects when a person speaks on the scene and the corresponding transcription. In addition, to show the complete process of preparing AnnoTheia for a language of interest, we also describe the adaptation of a pre-trained model for active speaker detection to Spanish, using a database not initially conceived for this type of task. The AnnoTheia toolkit, tutorials, and pre-trained models are available at \url{https://github.com/joactr/AnnoTheia/}. 
 \\ \newline  
}

\begin{document}

\maketitleabstract

\section{Introduction}
\label{sec:introduction}


\begin{table*}[!htbp]
    \centering
    \footnotesize
    \addtolength{\tabcolsep}{-0.3em}
    \begin{adjustbox}{max width=\textwidth}
    \begin{threeparttable}
    \begin{tabular}{lccccccccccc}
         \toprule
         & \textbf{English} & \textbf{Persian} & \textbf{Spanish} & \textbf{French} & \textbf{Portuguese} & \textbf{Italian} & \textbf{Chinese} & \textbf{Russian} & \textbf{German} & \textbf{Greek} & \textbf{Arabic} \\ \midrule
         \textbf{MuAViC}\tnote{1} & 438.9\tnote{$^\dagger$} & - & 181.0 & 179.0 & 156.0 & 104.0 & - & 52.0 & 13.0 & 29.0 & 19.0 \\
         \textbf{CMU-MOSEAS}\tnote{2} & - & - & 16.3 & 15.6 & 16.0 & - & - & - & 18.6 & - \\
         \textbf{LRS2-BBC}\tnote{3}& 224.5 & - & - & - & - & - & - & - & - & - & - \\
         \textbf{LRS3-TED}\tnote{4} & 438.9 & - & - & - & - & - & - & - & - & - & - \\
         \textbf{CMLR}\tnote{5} & - & - & - & - & - & - & 86.5 & - & - & - & - \\
         \textbf{Arman-AV}\tnote{6} & - & 220.0 & - & - & - & - & - & - & - & - & - \\
         \textbf{LIP-RTVE}\tnote{7} & - & - & 13.0 & - & - & - & - & - & - & - & - \\ \midrule
         \textbf{Total of Hours} & 663.4 & 220.0 & 210.3 & 194.6 & 172.0 & 104.0 & 86.5 & 52.0 & 31.6 & 29.0 & 19.0 \\ \bottomrule
    \end{tabular}
    \begin{tablenotes}
        \footnotesize
        \item[1] \cite{anwar2023muavic}
        \item[2] \cite{zadeh2020moseas}
        \item[3] \cite{afouras2018deep}
        \item[4] \cite{afouras2018lrs3}
        \item[5] \cite{zhao19cmlr}
        \item[6] \cite{peymanfard2023multi}
        \item[7] \cite{lrec2022liprtve}
        \item[$\dagger$] The MuAViC database was sourced from TED and TEDx talks: thus, it includes directly the LRS3-TED corpus.
    \end{tablenotes}
    \caption{Language coverage in terms of hours by databases focused on unconstrained continuous audio-visual speech recognition. Corpora oriented to word-level classification, collected in controlled recording studios, or not providing transcriptions were not considered.}
    \label{tab:av-lang-cov}
    \end{threeparttable}
    \end{adjustbox}
\end{table*}

Speech technologies aim to develop tools capable of enabling human-machine interaction via different types of communication. Multiple research areas are included in this vast field, e.g., speech recognition \cite{baevski2020wav2vec,radford2023robust,shi2022learning,maja2021conformers}; speech synthesis \cite{shen2018natural,transformer2019tts,liu2023synthvsr}; speech translation \cite{jia19s2strans,lee2022direct,huang2023avspeechtrans}; sign language recognition \cite{koller2015sign,camgoz20cpvr,albanie2021bobsl}; silent speech interfaces \cite{denby2010silent,ssi2020review}; speech-based disease detection \cite{abad2022dementia,milling2022speech}; speech enhancement and separation \cite{pascual2017isegan,yen2023colddiff,huang2022sess}; speaker diarization \cite{fujita19diarization,bredin21segmentation}; spoken dialogue systems \cite{jokinen2022spoken}. All these studies reveal the importance that speech technologies and their applications can represent in our daily lives.

One of the most widely studied tasks in the field is automatic speech recognition. However, although more than 7,000 known languages are spoken around the world \cite{lewis2009ethnologue}, only a small fraction of them are currently covered by these speech technologies \cite{pratap2023scaling}. Taking into account that many of these languages are considered endangered \cite{bromham2022global}, the lack of support from current technologies may contribute negatively to this situation. For this reason, to avoid any type of discrimination or exclusion, different works have promoted scaling speech technologies to cover the greatest possible number of languages. Massive multi-lingual speech corpora collections \cite{ardila2020common,salesky21_interspeech,anwar2023muavic,pratap2023scaling}, large multi-lingual speech recognition models \cite{radford2023robust,pratap2023scaling,zhang2023google,mmasr23watanabe}, self-supervised speech representations \cite{baevski2020wav2vec,ravanelli2020pase,babu22xlsr,shi2022learning}, as well as the organization of challenges, such as the ML-SUPERB \cite{shi2023mlsuperb}, have been some of the most common approaches to alleviate this inequality. Nevertheless, despite all these efforts, most studies are still benchmarked on English \cite{pratap2023scaling}.

In addition, the described approaches mainly focused on auditory-based speech recognition, contrasting with the recent trend towards developing audio-visual speech technologies \cite{maja2021conformers,shi2022learning,liu2023synthvsr}. Tasks such as automatic lipreading \cite{afouras2018deep,ma2022visual,prajwal2022sub}, audio-visual speech recognition \cite{maja2021conformers,shi2022learning,burchi2023wacv}, or audio-visual speech synthesis \cite{liu2023synthvsr} have received increasing interest in the last decades. However, the scarcity of data for low-resource languages is exacerbated when this type of tasks is addressed due to the difficulty of collecting data where audio and video are aligned. Table \ref{tab:av-lang-cov} reflects the limited number of languages covered by the audio-visual corpora that have been publicly released for unconstrained audio-visual speech recognition. Corpora oriented to word-level classification \cite{chung2017lip,lrw2019chinese,egorov2021lrwr} or collected in controlled recording studios \cite{harte2015tcd,fernandez2017towards} were not considered, since they do not represent the actual nature of speech\footnote{A more comprehensive review of existing audio-visual databases can be found in \cite{fernandez2018survey}.}. 

A special case is the AV-Speech \cite{ephrat18avspeech}, a large-scale database offering over 4k hours of data for multiple languages, which was also not considered because it does not provide transcriptions. It should be noted that this corpus could be used for pre-training multilingual encoders in a self-supervised manner \cite{shi2022learning}. However, the current lack of audio-visual language resources would impede a more effective use of this technology \cite{anwar2023muavic}.

Despite all these efforts in the audio-visual domain, there is no comparison to large-scale audio-based speech databases that cover more than a thousand languages with thousands of hours of data \cite{pratap2023scaling,zhang2023google}. Besides, the inequality in terms of the number of hours among the different languages reflected in Table \ref{tab:av-lang-cov} and the fact that most studies based on audio-visual technologies are mainly benchmarked on English, highlights the need to promote and encourage the collection and annotation of new audio-visual databases to cover the largest possible number of languages.

For this reason, numerous works have described multi-step pipelines to collect and annotate their own audio-visual databases \cite{afouras2018deep,afouras2018lrs3,peymanfard2023multi,ephrat18avspeech}. However, while building these pipelines is not trivial, to the best of our knowledge, no toolkit has been publicly released for the research community. In addition, one of the most important components of these pipelines is the Active Speaker Detection (ASD) module \cite{tao2021talknetasd,liao2023lightasd,min2022spell}, which allows us to identify the person on the scene who is talking. Nevertheless, these ASD models are also heavily benchmarked on English \cite{tao2021talknetasd,liao2023lightasd,min2022spell}.

Motivated by this fact, we present AnnoTheia, a semi-automatic toolkit that detects when a person speaks on the scene and the corresponding transcription. One of the most notable aspects of the proposed toolkit is the flexibility to replace a module with another of our preference or, where appropriate, adapted to our language of interest. Therefore, to show the complete process of preparing AnnoTheia for a language of interest, we also describe in this paper the adaptation of a pre-trained ASD model to Spanish, using a database not initially conceived for this type of task.

These were the main reasons that motivated our work, whose key contributions are: 

\begin{itemize}
    \item We present AnnoTheia, a semi-automatic annotation toolkit to encourage the largest possible language coverage for audio-visual speech technologies, highlighting its intuitive user interface and its flexibility to replace the different modules that composed it. 

    \item We described how to adapt a pre-trained ASD model to a language of interest, even in those occasions where there are no databases originally conceived for this type of task.
    
    \item We consequently present RTVE-ASD, a new corpus for ASD in Spanish collected from TV newscast programs, where multiple challenging in-the-wild situations can be found.

    
\end{itemize}


\begin{figure*}[htbp]
  \centering
  \includegraphics[width=\textwidth]{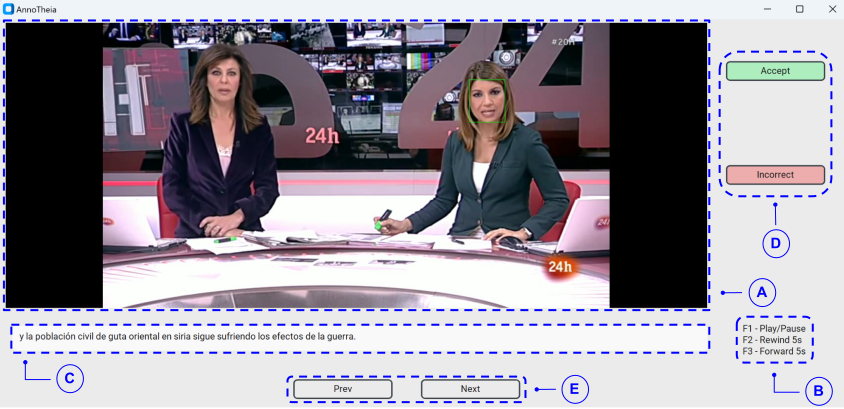}
  \caption{AnnoTheia toolkit user interface. {\color{blue} \textbf{A}}: Video display of the scene candidate to be a new sample of the future database. An overlying green bounding box highlights the active speaker detected by the toolkit. {\color{blue} \textbf{B}}: Keyword legend to control the video display. {\color{blue} \textbf{C}}: Transcription automatically generated by the toolkit. It can be edited by the annotator. {\color{blue} \textbf{D}}: Buttons to allow the annotator to accept or discard the candidate scene sample. {\color{blue} \textbf{E}}: Navigation buttons through candidate scenes. It can be useful to correct possible annotation mistakes.}
  \label{fig:toolkit-user-interface}
\end{figure*}

\section{The AnnoTheia Toolkit}
\label{sec:toolkit}
We present AnnoTheia, the first open-source toolkit for the semi-automatic annotation of audio-visual speech resources, promoting the further development of speech technologies for low-resource languages. In this section, we describe the multiple modules that compose the toolkit, as well as the design of the user interface. However, we highlight the flexibility with which the toolkit was provided when replacing one of these modules with another of our preference or, as described in Section \ref{sec:asd}, adapted to our language of interest.

\medskip

\noindent \textbf{Scene Bounding Detection.} PySceneDetection\footnote{\url{https://github.com/Breakthrough/PySceneDetect}} is a toolkit designed to detect bounding between scenes in videos. Specifically, it computes a score representing the difference in content between the current and previous frame in the HSV color space for each video frame. In this way, we were able to accelerate the processing of large video clips by checking if at least one person was present on the scene simply by checking the first frame that composed each detected scene. Hence, using the face detector described below, we discarded those scenes where no people were present.

\noindent \textbf{Face Detection.} Using Dual Shot Face Detector \cite{li2019dualshot}, we were able to accurately extract faces from the scenes previously selected. However, more than one person could appear on the scene at the same time, hindering the subsequent Active Speaker Detection. Therefore, we implemented a frame-by-frame face-distance comparison algorithm, where each face was matched to a specific individual based on proximity, allowing for the precise assignment of faces to their respective individuals throughout the video sequence. The detected bounding boxes were saved not only because we needed the cropped faces for the Active Speaker Detection module described below, but also because this information might be useful once the future database is compiled. An alternative would be using the state-of-the-art RetinaFace face detector \cite{deng2020retina}.

\noindent \textbf{Face Alignment.} On multiple occasions, it can also be useful to consider a specific region of the face, as it is the case with the visual speech recognition task, also known as automatic lipreading \cite{afouras2018deep,ma2022visual,shi2022learning}. For this reason, we also introduced a face alignment module based on the Face Alignment Network architecture proposed by \newcite{bulat2017facealign}. Hence, our toolkit will also provide 68 2-dimensional facial landmarks for each frame of the detected scene. An alternative would be using the face landmark detector provided by the Google's MediaPipe framework\footnote{\url{https://developers.google.com/mediapipe/solutions/vision/face_landmarker}}.

\noindent \textbf{Active Speaker Detection.} Once the different people appearing on the scene were detected, we used the TalkNet-ASD model \cite{tao2021talknetasd} to identify who of them was really speaking, since, on certain occasions, we may encounter dubbing or voiceovers. Besides, we studied different strategies when processing large video clips, concluding that a non-overlapping window-slicing method offered the best trade-off between quality and speed. A more detailed description of the TalkNet-ASD model architecture can be found in Subsection \ref{sec:talknet}. An alternative would be using the state-of-the-art Light-ASD model \cite{liao2023lightasd}.

\noindent \textbf{Post-Process.} The ASD module previously described is not exempt from errors. Therefore, we applied a smoothing average strategy for stabilizing the complete scene detection process. This method entailed calculating the mean score provided by the ASD model for a defined sliding window of frames, assigning this averaged score to the central frame within the window. The window then shifts by one frame at a time, repeating the process across the entire sequence. This approach effectively mitigates potential fluctuations in classification scores by incorporating information from neighboring frames, leading to a more consistent and stable classification.

\noindent \textbf{Scene Trimming.} We only considered as candidate scene samples those that encompassed frames in which the ASD module detected that a person was actively talking, trimming the scenes with a margin of a small number of frames both at the beginning and at the end of the scene.

\noindent \textbf{Automatic Transcription.} Whisper is a multi-lingual speech recognition system covering around 100 different languages \cite{radford2023robust}. By applying this technology to the raw waveform of our trimmed scenes, we were able to obtain word-level aligned transcriptions, thus facilitating the annotator's task. Although the model automatically detects the spoken language, it can also be manually declared by the user. An alternative would be using the state-of-the-art MMS models \cite{pratap2023scaling}\footnote{\url{https://github.com/facebookresearch/fairseq/tree/main/examples/mms}}.

\noindent \textbf{User Interface.} In order to facilitate the annotator's task, an intuitive user interface has been designed to accelerate the data collection process. Details of the design can be found in Figure \ref{fig:toolkit-user-interface}.

\section{Fine-Tuning an ASD Model}
\label{sec:asd}

This section describes the entire process we followed to prepare AnnoTheia for a language of interest by fine-tuning a pre-trained ASD model even in those situations where there are no databases originally conceived for this type of task. In our case, we adapted the TalkNet-ASD model \cite{tao2021talknetasd} pre-trained for the AVA-ActiveSpeaker benchmark \cite{roth2020ava} to the Spanish language using the LIP-RTVE database \cite{lrec2022liprtve}. Hence, in this section, we described the model architecture, how to construct a dataset to estimate an ASD model, and the training setup, as well as a discussion of the findings we observed in our results.

\subsection{The TalkNet-ASD Model}
\label{sec:talknet}

\begin{figure*}[htbp]
  \centering
  \includegraphics[width=0.9\textwidth]{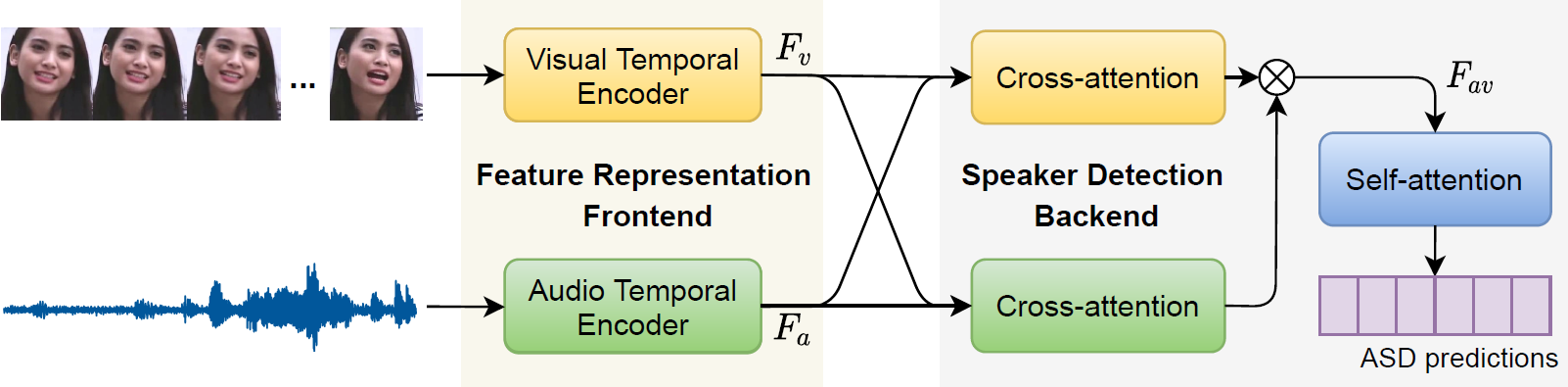}
  \caption{The overall architecture of the TalkNet-ASD model.}
  \label{fig:talknet}
\end{figure*}

The TalkNet-ASD model \cite{tao2021talknetasd} was explicitly designed for the ASD task, where the model aims to identify who of the people appearing in a video clip is actually speaking and at what time. It required two simultaneous inputs: raw grayscale frames of the cropped person's face, providing the visual cues, and pre-processed audio features in the form of 13 Mel Frequency Cepstral Coefficients (MFCC) \cite{gales2008application}, representing the audio content. It was able to achieve state-of-the-art results on the AVA-ActiveSpeaker benchmark \cite{roth2020ava}. As reflected in Figure \ref{fig:talknet}, the TalkNet-ASD architecture is composed of three main modules.

\noindent \textbf{Window Sampling.} Due to the unfeasibility, in computational terms, of processing large videos in their entirety, TalkNet-ASD worked at the window sample level. In the original work, the authors studied multiple window sizes, covering contexts of different numbers of seconds. 

\noindent \textbf{Feature Representation Frontend.} Once the context windows are sampled, the visual and acoustic cues described above are each processed by an independent encoder.
\begin{itemize}
    \item \textbf{Visual Temporal Encoder.} This encoder was aimed to capture long-term facial expression dynamics. It comprises a visual frontend for spatial information and a visual temporal block for temporal patterns. The frontend consists of a 3D convolutional layer followed by a 2D ResNet-18 \cite{resnet}, providing a sequence of frame-based embeddings. Subsequently, a block mainly based on depth-wise separable convolutional layers \cite{separable-conv} with residual connections is used to model temporal relationships of the previous visual representation.

    \item \textbf{Audio Temporal Encoder.} By using a 2D ResNet-34 \cite{resnet} with a squeeze-and-excitation module \cite{hu2018squeeze}, this encoder was able to learn audio representations from temporal dynamics. Given the MFCCs described above, the encoder produces a sequence of audio embeddings that, thanks to the use of dilated convolutions when defining the ResNet-34, match the length of the visual embedding sequence.
    
\end{itemize}

\noindent \textbf{Speaker Detection Backend.} This module is composed of two network blocks mainly based on attention mechanisms \cite{vaswani2017attention}. First, a cross-attention network is employed to dynamically capture audio-visual interactions by aligning the audio and visual embeddings previously obtained by the feature frontend. Once a joint audio-visual representation is computed, a self-attention network focuses on temporal relationships, distinguishing whether the person is speaking or not across the video frames.

\noindent \textbf{Post-Process} Another important aspect to highlight is the use of an optimum classification threshold. Because the main purpose of the proposed toolkit is to provide the annotator with candidate scene samples to create a new database, we were interested in losing, within reason, as few candidate scenes as possible due to false negatives that the model could predict, delegating the final decision to the annotator's criteria. Therefore, we computed the optimum threshold\footnote{\url{https://machinelearningmastery.com/threshold-moving-for-imbalanced-classification/}} to find the best trade-off between the true-positive and false-positive rates on a validation set.

\subsection{The ASD-RTVE corpus}

This section describes how, from a database that was not originally conceived to address this type of tasks, we were able to construct a dataset to estimate and evaluate an ASD model.

We present ASD-RTVE, a new corpus for ASD in Spanish. It was constructed using the LIP-RTVE database \cite{lrec2022liprtve}, where multiple challenging in-the-wild situations can be found, as it was collected from TV newscast programs. Due to the fact that LIP-RTVE was conceived to address audio-visual speech recognition, all their samples include a person talking during the entire video clip. Therefore, in order to estimate a model focused on the ASD task, we defined different types of samples, intending to cover all the possible situations we might find when applying our model in a realistic application scenario.

\noindent \textbf{Positive Samples.} We consider a positive sample when audio and video correspond to each other, i.e., the voice is aligned with the talking face.

\noindent \textbf{Negative Samples.} With the intention of providing the model with a certain robustness against all types of undesired situations that might occur in a realistic application scenario, we defined three different types of negative samples:

\begin{itemize}
    \item \textbf{Temporal Mismatch.} The audio and video cues correspond to each other. However, the audio was randomly shifted to simulate a temporal mismatch, as long as the overlap between both modalities did not exceed 50\%. This type of sample is aimed to make the model discard those scenes where audio and video cues are not aligned.
    
    \item \textbf{Partial Speaker Mismatch.} In this case, although the voice and the face belong to the same person, the audio and video cues do not correspond to each other.   This type of sample is aimed to ensure that the model does not learn to simply associate a voice with its corresponding face.
    
    \item \textbf{Complete Speaker Mismatch.} The voice and the face belong to different people. This type of sample was aimed to provide the model examples of situations we can find in a real scenario, such as dubbing or voiceovers.

\end{itemize}

Specifically, we split the corpus into training, validation, and test sets, ensuring a balance between all types of samples previously described. More details about the data sets defined for the ASD-RTVE database can be found in Table \ref{tab:asd-rtve}.

\begin{table}[!htbp]
    \centering

    \begin{adjustbox}{width=0.9\columnwidth}
    \begin{tabular}{lcccc}
        \toprule
        \multirow{2}{*}[-2.5pt]{\textbf{Dataset}} & \multicolumn{3}{c}{\textbf{No. of Speakers}} & \multirow{2}{*}[-2.5pt]{\textbf{\begin{tabular}[c]{@{}c@{}}No. of\\Utterances\end{tabular}}} \\
        \cmidrule{2-4}
        \textbf{} & \textbf{Female} & \textbf{Male} & \textbf{Total} &\\
        \midrule
        \textbf{Traning} & 19 & 10 & 29 & 100k\\
        \textbf{Validation} & 65 & 86 & 151 & 30k\\
        \textbf{Test} & 76 & 67 & 143 & 30k\\
        \midrule
        \textbf{Total} & 160 & 163 & 323 & 160k\\
        \bottomrule
    \end{tabular}
    \end{adjustbox}
    \caption{Details of the ASD-RTVE corpus.}
    \label{tab:asd-rtve}
\end{table}

It should be noted that, if we do not have a database from which to generate data to train an ASD model for our language of interest, we can always use public sources to compile it, e.g., a set of vlogs where a single person is usually talking in front of the camera during the entire video clip can be used to obtain raw audio-visual data, and then, by following, the procedure described above, the ASD-oriented dataset could be created. In the worst-case scenario, an alternative approach would be to use an ASD module trained for English, for instance, and then set its decision threshold to 0.0. We are aware that this may entail more effort from the annotator's point of view, but we still believe the toolkit can speed up the process.

\subsection{Implementation Details}
Experiments were conducted on a GeForce RTX 4090 GPU with 24GB memory. The design of our ASD model was based on the official implementation\footnote{\url{https://github.com/TaoRuijie/TalkNet-ASD}} of the TalkNet-ASD model.

\noindent \textbf{Training.} In all our experiments, we used the Adam optimizer \cite{kingma2014adam} with a learning rate of 0.0001 decaying through a linear scheduler across 9 epochs, with a batch size of 32 samples. Because we fine-tuned the TalkNet-ASD model pre-trained for the AVA-ActiveSpeaker database \cite{roth2020ava}, its original architecture was not modified, referring readers to \newcite{tao2021talknetasd} to a more detailed description. Nonetheless, for the rest of the hyper-parameters, multiple settings were explored until we found the optimum we described above.

\noindent \textbf{Evaluation.} Similar to other works in the literature \cite{tao2021talknetasd,liao2023lightasd,min2022spell}, we reported our results in terms of different evaluation metrics, namely the accuracy, the mean Average Precision (mAP), and the Area Under the Curve (AUC).

\subsection{Results \& Discussion}

Similar to \newcite{tao2021talknetasd}, we studied how the size of the context window was affecting our system performance. Table \ref{tab:asd-window-ablation} illustrates a clear positive correlation on the validations set. As the number of context seconds increased, the ASD model showed a consistent upward trend, suggesting that incorporating more contextual information leads to more accurate and reliable predictions regarding speech activity detection.

\begin{table}[!htbp]
    \centering

    \begin{adjustbox}{width=\columnwidth}
    \begin{tabular}{ccccc}
        \toprule
        \multicolumn{2}{c}{\textbf{Context Window}} & \multirow{2}{*}{\textbf{Accuracy}} & \multirow{2}{*}{\textbf{mAP (\%)}} & \multirow{2}{*}{\textbf{AUC (\%)}} \\
        \cmidrule{1-2}
        {\small no. of frames} & {\small no. of seconds} &  &  &  \\
         \midrule
         \textbf{5} & \textbf{0.20} & 70.2{\scriptsize$\pm$0.5} & 60.0 & 75.3 \\
         
         \textbf{9} & \textbf{0.36} & 79.2{\scriptsize$\pm$0.5} & 72.0 & 84.2 \\
         
         \textbf{13} & \textbf{0.52} & 83.5{\scriptsize$\pm$0.4} & 77.7 & 88.0 \\
         
         \textbf{17} & \textbf{0.68} & 85.7{\scriptsize$\pm$0.4} & 78.4 & 91.2 \\
         
         \textbf{21} & \textbf{0.84} & 89.9{\scriptsize$\pm$0.3} & 85.2 & 94.8 \\
         
         \textbf{25} & \textbf{1.00} & 91.8{\scriptsize$\pm$0.3} & 87.5 & 95.8 \\
         
         \textbf{35} & \textbf{1.40} & 93.1{\scriptsize$\pm$0.3} & 90.7 & 98.5 \\
         
         \textbf{43} & \textbf{1.72} & 97.0{\scriptsize$\pm$0.2} & 94.8 & 99.3 \\
         
         \textbf{51} & \textbf{2.04} & 97.3{\scriptsize$\pm$0.2} & 95.8 & 99.5 \\
         \bottomrule
    \end{tabular}
    \end{adjustbox}
    \caption{Results for the ASD-RTVE validation set depending on the context window size.}
    \label{tab:asd-window-ablation}
\end{table}

Furthermore, we evaluated the ASD models estimated for each window size in terms of inference time. However, no significant differences were found between the studied context windows. Therefore, for AnnoTheia, we used the ASD model trained with 51-frame windows, which was able to achieve, on the ASD-RTVE test set, results around 95.4{\small $\pm$0.2} accuracy, 91.6 \% mAP, and 99.3 \% AUC.


\section{Conclusions \& Future Work}
\label{sec:conclusion}
In this work, we presented the AnnoTheia toolkit to facilitate the task of annotating audio-visual resources for speech technologies, thus promoting further research on low-resource languages in the field. In addition, because some of its modules are language-dependent, we showed the complete process of preparing AnnoTheia for a language of interest, even in those occasions where there are no databases originally conceived for that purpose. Regarding our future work, we are considering introducing further refinements to improve the toolkit both from a performance and usability perspective. One important contribution would be integrating a module capable of ensuring speaker-independent partitions, e.g., via face, body, and voice clustering \cite{brown2021iccv,deng2019arcface}. Furthermore, we plan to conduct a comprehensive experiment to properly assess the effectiveness and advantages of using the AnnoTheia toolkit versus a completely manual procedure.

\section{Limitations}
\label{sec:limitations}
Our toolkit's main limitation is the need to adapt the ASD module to the language of interest, which could hinder the experience for users unfamiliar with training machine learning-based systems. Another aspect to consider is the possibility that the pre-trained models composing the toolkit can be biased. Due to the lack of demographically diverse data, certain social groups could be underrepresented because of their age, gender, and cultural background, among other factors.

\section{Ethical Considerations}
\label{sec:ethical}
The toolkit we present can be used to collect and annotate audio-visual databases from social media where a person's face and voice are recorded. Both information cues could be considered as biometrics \cite{nagrani17voxceleb}, raising privacy concerns regarding a person's identity. Therefore, any data collection must protect privacy rights, taking necessary measures and always asking all the people involved for permission beforehand.

\section{Acknowledgements}
\label{sec:acknowledgements}
The work of José-M. Acosta-Triana was in the framework of the Valencian Graduate School and Research Network for Artificial Intelligence (ValgrAI) funded by Generalitat Valenciana. The work of David Gimeno-Gómez and Carlos-D. Martínez-Hinarejos was partially supported by Grant CIACIF/2021/295 funded by Generalitat Valenciana and by Grant PID2021-124719OB-I00 under project LLEER (PID2021-124719OB-100) funded by MCIN/AEI/10.13039/501100011033/ and by ERDF, EU A way of making Europe.

\section{Bibliographical References}
\label{sec:bibliographical}
\bibliographystyle{lrec2022-bib}
\bibliography{lrec2022-example}

\end{document}